\documentclass{article}

\usepackage{amsmath}
\usepackage{amsfonts}
\usepackage{amssymb}
\usepackage{graphicx}
\usepackage{url}
\usepackage[authoryear]{natbib}

\usepackage{appendix}

\usepackage{graphicx}		
\usepackage{float}			
\usepackage{subfloat}		
\usepackage{subfigure}		
\usepackage{lscape}			

\usepackage{enumitem}

\usepackage{gensymb}
\usepackage{amsmath}
\usepackage{amssymb}
\usepackage{amsthm}
\usepackage{exscale}
\usepackage{textcomp}		

\usepackage{algorithmic}
\usepackage{algorithm}

\usepackage{array}
\usepackage{tabulary}
\usepackage{multirow}
\usepackage[table]{xcolor}
\usepackage{ctable}
\usepackage{booktabs}		

\usepackage{url}
\usepackage{hyperref}

\newcommand{\link}[1]{\href{#1}{\textnormal{#1}}}

\usepackage{listings}

\usepackage{titlesec}

\usepackage{enumitem}

\title{PyTorch-based Geometric Learning with Non-CUDA Processing Units:\\Experiences from Intel Gaudi-v2 HPUs}
\author{
Fanchen Bu and Kijung Shin\\
KAIST \\
\{boqvezen97, kijungs\}@kaist.ac.kr\\
}

\date{}

\begin{document}
\maketitle

\begin{abstract}
Geometric learning has emerged as a powerful paradigm for modeling non-Euclidean data, especially graph-structured ones, with applications spanning social networks, molecular structures, knowledge graphs, and recommender systems.
While Nvidia's CUDA‑enabled graphics processing units (GPUs) largely dominate the hardware landscape, emerging accelerators such as Intel's Gaudi Habana Processing Units (HPUs) offer competitive performance and energy efficiency.
However, the usage of such non-CUDA processing units requires significant engineering effort and novel software adaptations.
In this work, we present our experiences porting PyTorch‑based geometric learning frameworks to Gaudi‑v2 HPUs.
We introduce a collection of core utilities that restore essential operations (e.g., scatter, sparse indexing, $k$‑nearest neighbors) on Gaudi‑v2 HPUs, and we consolidate sixteen guided tutorials and eleven real‑world examples with diagnostic analyses of encountered failures and detailed workarounds.
We collect all our experiences into a publicly accessible GitHub repository.
Our contributions lower the barrier for researchers to experiment with geometric-learning algorithms and models on non‑CUDA hardware, providing a foundation for further optimization and cross‑platform portability.
\end{abstract}

\section{Introduction}
Graphs are a fundamental representation for a wide range of real‑world systems, including social networks~\cite{nettleton2013data}, molecular structures~\cite{trinajstic2018chemical}, knowledge graphs~\cite{hogan2021knowledge}, and recommender systems~\cite{wu2022graph}, all of which naturally take the form of nodes connected by edges.

Geometric learning, which generalizes deep learning to non‑Euclidean domains~\cite{bronstein2017geometric}, especially on graph-structured data~\cite{xia2021graph,wu2020comprehensive}, has emerged as a powerful paradigm for capturing relational and structural information.
Graph neural network (GNN) models such as graph convolutional networks (GCNs)~\cite{kipf2016semi}, graph attention networks (GATs)~\cite{velivckovic2017graph,brody2021attentive,Lee2023AEROGNN}, and GraphSAGE~\cite{hamilton2017inductive} have demonstrated success across tasks from node classification~\cite{xiao2022graph} to link prediction~\cite{zhang2018link} and recommendation~\cite{lee2024revisiting}.

While Nvidia's CUDA‑enabled graphics processing units (GPUs) have become the de facto standard for accelerating deep learning workloads, there is a growing ecosystem of alternative processing units that offer different trade‑offs in cost, power efficiency, and architectural specialization~\cite{lee2024debunking}.
Non‑CUDA processing units, such as Google's tensor processing unit (TPUs) and Intel's Gaudi HPUs, are gaining traction in both industry and research. Supporting geometric learning workloads on these non-CUDA hardware platforms is essential to broaden hardware choice, optimize for diverse deployment scenarios, and leverage specialized hardware capabilities.

Intel’s Gaudi HPUs exemplify non-CUDA accelerators tailored for deep learning.
Gaudi HPUs have features like high-throughput matrix engines, programmable tensor cores, and a software stack compatible with Pytorch,\footnote{\link{https://pytorch.org}} a widely used open-source deep learning framework.
Gaudi HPUs aim to deliver competitive performance and efficiency for large-scale training.
However, adapting geometric-learning models and libraries originally developed for CUDA presents challenges, including unsupported sparse operations and issues with backpropagation in key kernels.

In this paper, we share our experiences adapting PyTorch-based geometric learning workflows to Intel's Gaudi‑v2 HPUs.
Our implementations, tutorials, examples, and diagnostic analyses are consolidated into a publicly accessible GitHub repository.\footnote{\link{https://github.com/NAVER-INTEL-Co-Lab/gaudi-geometric-learning}}
This repository includes Gaudi-compatible reimplementations of core libraries, practical workarounds for unsupported features, and detailed guidance on environment setup and debugging.
By openly sharing these resources, we aim to lower the entry barrier for researchers and practitioners interested in exploring graph neural networks on non-CUDA hardware platforms.

\section{Repository Overview}
Our GitHub repository contains three main parts:
\begin{itemize} 
    \item \textbf{Documentation}: A user-friendly README that includes environment setup instructions, guidance for launching jobs on Gaudi, and links to relevant external materials.
    \item \textbf{Core Utilities}: A suite of helper modules designed to bridge gaps in existing libraries, enabling operations such as scatter, sparse indexing, and neighborhood computations on Gaudi without requiring low-level programming.
    \item \textbf{Tutorials and Examples}: A collection of step-by-step tutorials and real-world examples, covering tasks from basic graph convolutional network (GCN) construction to link prediction and point-cloud classification. Each tutorial and example is accompanied by runnable code and analyses of challenges encountered on Gaudi-v2.
\end{itemize}

\section{Core Utilities}
In this section, we describe the core utilities we developed to fill functional gaps in PyTorch-based geometric learning on Gaudi‑v2.

\subsection{Scatter Operations}
Scatter operations are fundamental in geometric learning for aggregating information from neighboring nodes.
The \texttt{torch-scatter} library provides optimized scatter operations,\footnote{\link{https://github.com/rusty1s/pytorch\_scatter}} which are widely used in geometric learning implementations.

\begin{figure}[t]
\centering
\includegraphics[width=0.5\textwidth]{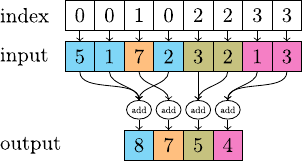}
\caption{Example of the scatter operation with the sum reduction (\texttt{scatter\_add}), which is adapted from \texttt{torch-scatter} documentation (\link{https://pytorch-scatter.readthedocs.io}) and used under the MIT License.}
\label{fig:scatter}
\end{figure}

See Figure~\ref{fig:scatter} for an example of the scatter operation with the sum reduction.
In this example, the \texttt{input} tensor is aggregated into the \texttt{output} tensor according to the \texttt{index} tensor.
Each value in \texttt{input} is added to the position in \texttt{output} specified by the corresponding value in \texttt{index}.
In other words, the values in \texttt{input} corresponding to the same \texttt{index} are grouped together.

\textbf{Challenge:} The \texttt{torch-scatter} library offers implementations for CPU and CUDA backends but lacks native support for Gaudi HPUs, hindering its direct usage on this platform.

\textbf{Solution:} We re-implement scatter operations using pure PyTorch functions, ensuring compatibility with Gaudi-v2 HPUs.

\subsection{Sparse Operations}

Sparse tensors efficiently represent data with a high proportion of zero elements, reducing memory usage and computational overhead.
The \texttt{torch-sparse} library provides optimized sparse operations,\footnote{\link{https://github.com/rusty1s/pytorch\_sparse}} which are widely used in geometric deep learning and recommender systems to handle large sparse interaction graphs.

\textbf{Challenge:} The \texttt{torch-sparse} library offers implementations for CPU and CUDA backends but lacks native support for Gaudi HPUs, hindering its direct usage on this platform.

\textbf{Solution:} We transform several sparse tensor operations into equivalent scatter operations, ensuring compatibility with Gaudi-v2 HPUs.
For example, the multiplication between a sparse matrix and a dense matrix can be re-implemented via scatter-based aggregation, as shown in Algorithm~\ref{alg:sparse-scatter}.

\begin{algorithm}[t] 
\caption{Sparse-dense matrix multiplication via scatter} 
\label{alg:sparse-scatter}
\begin{algorithmic}[1] 
    \STATE \textbf{Input:} Sparse matrix $\mathbf{A}$ represented by indices ($\mathbf{u}, \mathbf{v}$) and values $\mathbf{val}$; dense matrix $\mathbf{X}$ 
    \STATE \textbf{Output:} Dense matrix result of multiplication $\mathbf{Y} = \mathbf{A}\mathbf{X}$ 
    \STATE Select rows from dense matrix: $\mathbf{X}_{\text{selected}} \leftarrow \mathbf{X}[\mathbf{v}]$ 
    \STATE Compute weighted rows: $\mathbf{X}_{\text{weighted}} \leftarrow \mathbf{val}[:,\text{None}] \odot \mathbf{X}_{\text{selected}}$
    \STATE Aggregate using scatter: $\mathbf{Y} \leftarrow \text{scatter\_add}(\mathbf{X}_{\text{weighted}}, \mathbf{u})$ 
    \RETURN $\mathbf{Y}$
\end{algorithmic} 
\end{algorithm}

\subsection{Advanced Graph Operations}

Advanced graph operations, such as $k$-nearest neighbors ($k$-NN) search and METIS partitioning, are integral to geometric learning workflows.
Libraries like \texttt{torch-cluster} provide these utilities,\footnote{\link{https://github.com/rusty1s/pytorch\_cluster}} often with CPU and CUDA-accelerated implementations.

\textbf{Challenge:} These graph utilities lack support for Gaudi HPUs.

\textbf{Solution:} We provide guidelines and implementations for executing these graph utilities on CPUs, followed by transferring the results to Gaudi HPUs for further processing. 

\begin{figure}[t!]
    \centering
    \vspace{2pt}
    \includegraphics[width=0.9\linewidth]{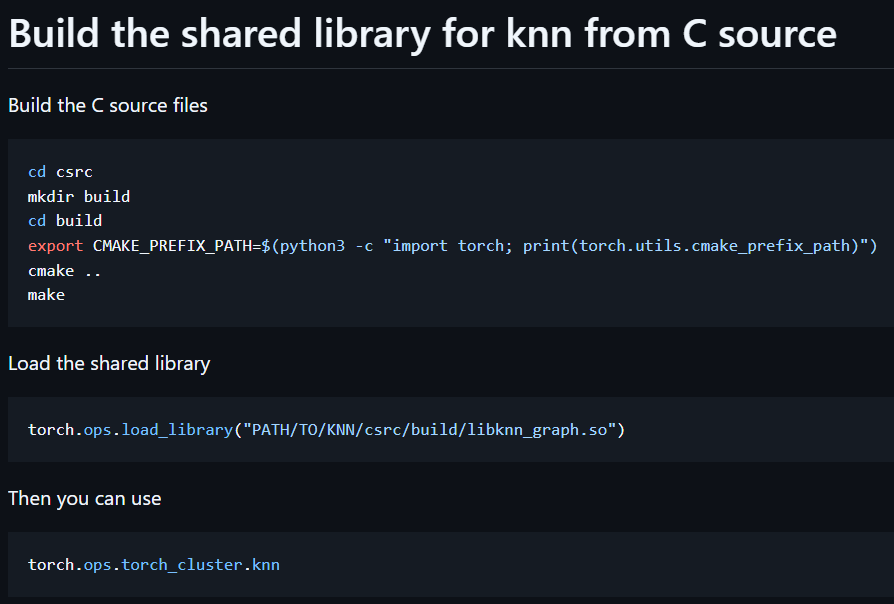}
    \caption{Example of building and loading the $k$-NN shared library from CPU-based C source code, enabling its use alongside Gaudi HPUs. Screenshot taken from our GitHub repository.}
    \label{fig:knn}
\end{figure}

See Figure~\ref{fig:knn}, where we show how to compile and load a custom implementation of the $k$-NN utility from its CPU-based source code, enabling it to be used seamlessly alongside Gaudi HPUs.

\section{Tutorials and Examples}
In this section, we introduce our collection of hands-on tutorials and examples.
Each tutorial and example is accompanied by executable code, detailed diagnostic analyses, and practical solutions addressing Gaudi-specific challenges.
We have adapted these materials from popular, established sources, including official PyTorch Geometric (PyG) tutorials and examples,\footnote{\link{https://pytorch-geometric.readthedocs.io/en/stable/get\_started/colabs.html}} to ensure broad relevance and accessibility.

\subsection{Tutorials}

The tutorials cover a wide range of complexity to facilitate incremental learning for users new to geometric learning:
\begin{itemize}
    \item \textbf{GCN and GAT Pipelines}: Fundamental tutorials that cover essential graph convolutional and attention mechanisms.
    \item \textbf{Embedding and Aggregation Functions}: Detailed guides illustrating how node embeddings and aggregation functions are implemented and executed efficiently on Gaudi.
    \item \textbf{Graph Pooling and Readout}: Tutorials demonstrating various techniques for graph-level representation learning and summarization.
    \item \textbf{DeepWalk and Node2Vec Practices}: Practical tutorials introducing representation learning via random walks, emphasizing node embeddings for downstream tasks.
\end{itemize}

\subsection{Examples}

We also include examples with realistic use cases:

\begin{itemize}
    \item \textbf{Node Classification}: Step-by-step workflows using benchmark datasets, including training and evaluation procedures.
    \item \textbf{Link Prediction}: Demonstrations of how to model relationships between entities using GNNs, including practical workflows for tasks such as recommendation.
    \item \textbf{Point-cloud Classification with DGCNN}: Examples illustrating geometric deep learning beyond graphs, specifically targeting 3D point-cloud classification tasks.
\end{itemize}

\begin{figure}[t!]
    \centering
    \includegraphics[width=0.9\linewidth]{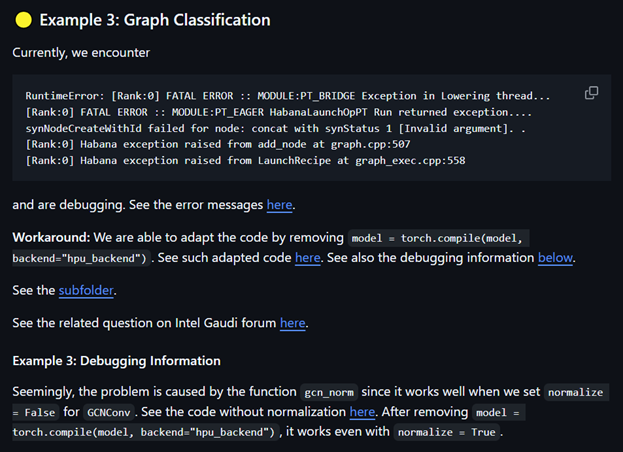}
    \caption{Example of workaround and debugging information for compatibility issues when adapting geometric learning code to Gaudi HPUs. Screenshot taken from our GitHub repository.}
    \label{fig:enter-label}
\end{figure}

\subsection{Diagnostic Analyses}

Adapting existing PyTorch Geometric tutorials and examples to Gaudi-v2 HPUs involved resolving a range of compatibility issues arising from unsupported operations, unexpected runtime errors, and subtle numerical mismatches.
To assist users in troubleshooting similar issues, we systematically documented our debugging process throughout the adaptation effort.
Figure~\ref{fig:enter-label} shows an example of such diagnostic documentation.

\section{Conclusion}
In this report, we have presented our efforts to adapt PyTorch‑based geometric learning workflows to the Intel Gaudi‑v2 HPUs.
By developing a suite of core utilities, we bridged critical functional gaps, making them seamlessly compatible with Gaudi. Our combined collection of step‑by‑step tutorials and real‑world examples not only demonstrates end‑to‑end GNN training on non‑CUDA hardware but also transparently shares the issues we encountered and the pragmatic workarounds that resolved them.
All of these resources are consolidated in a single, publicly accessible GitHub repository, lowering the barrier for practitioners to explore graph neural networks on alternative accelerators.


\section*{Acknowledgments}
This research was supported in part by the
NAVER-Intel Co-Lab.

\bibliographystyle{plainnat}
\bibliography{references}

\end{document}